\pdfoutput=1
\documentclass[11pt]{article}

\usepackage{ACL2023}

\usepackage{times}
\usepackage{latexsym}
\usepackage{graphicx}

\usepackage[T1]{fontenc}
\usepackage[utf8]{inputenc}

\usepackage{microtype}

\usepackage{inconsolata}

\usepackage{color}
\usepackage{tabularx}
\usepackage{inconsolata}
\usepackage{times}
\usepackage{amsmath}
\usepackage{amssymb}
\usepackage{subfigure}
\usepackage{multirow}
\usepackage{times}
\usepackage{latexsym}
\usepackage{graphicx}
\usepackage{booktabs}
\usepackage{array}
\title{From Complex to Simple: Unraveling the Cognitive Tree for Reasoning with Small Language Models}

\author{Junbing Yan$^1$, Chengyu Wang$^2$, Taolin Zhang$^2$, Xiaofeng He$^1$, \\
{\bf Jun Huang$^2$, Wei Zhang$^{1,3}$\thanks{\ \ \ Correspondence to Wei Zhang.}} \\
$^1$ School of Computer Science and Technology, East China Normal University \\ 
$^2$ Alibaba Group \\
$^3$ Shanghai Institute for AI Education \\
\texttt{\{junbingyan531,zhangwei.thu2011\}@gmail.com, hexf@cs.ecnu.edu.cn} \\
\texttt{\{chengyu.wcy,zhangtaolin.ztl,huangjun.hj\}@alibaba-inc.com}\\
}

\begin{document}

\maketitle

\begin{abstract}
Reasoning is a distinctive human capacity, enabling us to address complex problems by breaking them down into a series of manageable cognitive steps. 
%This transformative process empowers us to solve initially insurmountable problems.
Yet, complex logical reasoning is still cumbersome for language models.
Based on the dual process theory in cognitive science, we are the first to unravel the cognitive reasoning abilities of language models.
% Our framework progressively constructs a \textbf{Cognitive Tree} (CogTree) through an iterative process, where its root node of this tree is the query, and the leaf nodes are simple questions that can be directly answered.
% % Specifically, we coordinate an \textbf{implicit extraction} module (as the intuitive system) with an \textbf{explicit reasoning} module (as the reflective system).
% Specifically, we coordinate the implicit extraction module (as the intuitive system), which rapidly generates multiple answers by leveraging in-context examples; and the explicit reasoning module (as the reflective system), which scores the answers through comparative learning to guide the intuitive system in its next step of generation.
Our framework employs an iterative methodology to construct a \textbf{Cognitive Tree} (CogTree). The root node of this tree represents the initial query, while the leaf nodes consist of straightforward questions that can be answered directly. 
This construction involves two main components: the implicit extraction module (referred to as the intuitive system) and the explicit reasoning module (referred to as the reflective system). The intuitive system rapidly generates multiple responses by utilizing in-context examples, while the reflective system scores these responses using comparative learning. The scores guide the intuitive system in its subsequent generation step.
Our experimental results on two popular and challenging reasoning tasks indicate that it is possible to achieve a performance level comparable to that of GPT-3.5 (with 175B parameters), using a significantly smaller language model that contains fewer parameters (<=7B) than \textbf{5\%} of GPT-3.5.
\footnote{The source code will be released in the EasyNLP framework~\cite{DBLP:conf/emnlp/WangQZLLWWHL22}. URL: \url{https://github.com/alibaba/EasyNLP}}
% \footnote{The source code and experiment details of this paper can be obtained from \url{https://github.com/EdmundYanJ/CogTree}.}
\end{abstract}

\section{Introduction}

The human brain is akin to a garden, where instincts are seeds that sprout and grow, while reason acts as the gardener, pruning and nurturing the plants of knowledge to bloom into the flowers of enlightenment.
For machines, recently, Large Language Models (LLMs) have demonstrated their abilities to tackle diverse tasks through instantaneous question answering, exhibiting some levels of intelligence~\cite{instructGPT, zeroshotlearner,self-instruct}.

However, to cross the chasm between machines and humans, three main challenges still lie ahead: 
\textbf{1)} Reasoning ability. When it comes to mathematical and reasoning problems, the performance of LMs is still not satisfactory~\cite{Advances,mathword}. 
\textbf{2)} Cognition capacity. The evaluation and decision-making process regarding the problem and its current state is of paramount importance, especially when dealing with problems that involve lengthy reasoning chains or multi-step solutions. 
%complex mathematical questions requiring multi-step solutions. 
However, current methods~\cite{cot,tot} often lack comprehensive validation and tend to focus on verifying intermediate results~\cite{mathprompt}. 
\textbf{3)} Efficiency. The deployment and inference costs of LLMs are relatively high, especially when utilizing parameter-free inference enhancement techniques~\cite{cot,tot}. These techniques require extensive contexts and multiple steps of answer generation, leading to a further increase in inference costs and time.

\begin{figure}
\centering
\includegraphics[width=0.5\textwidth]{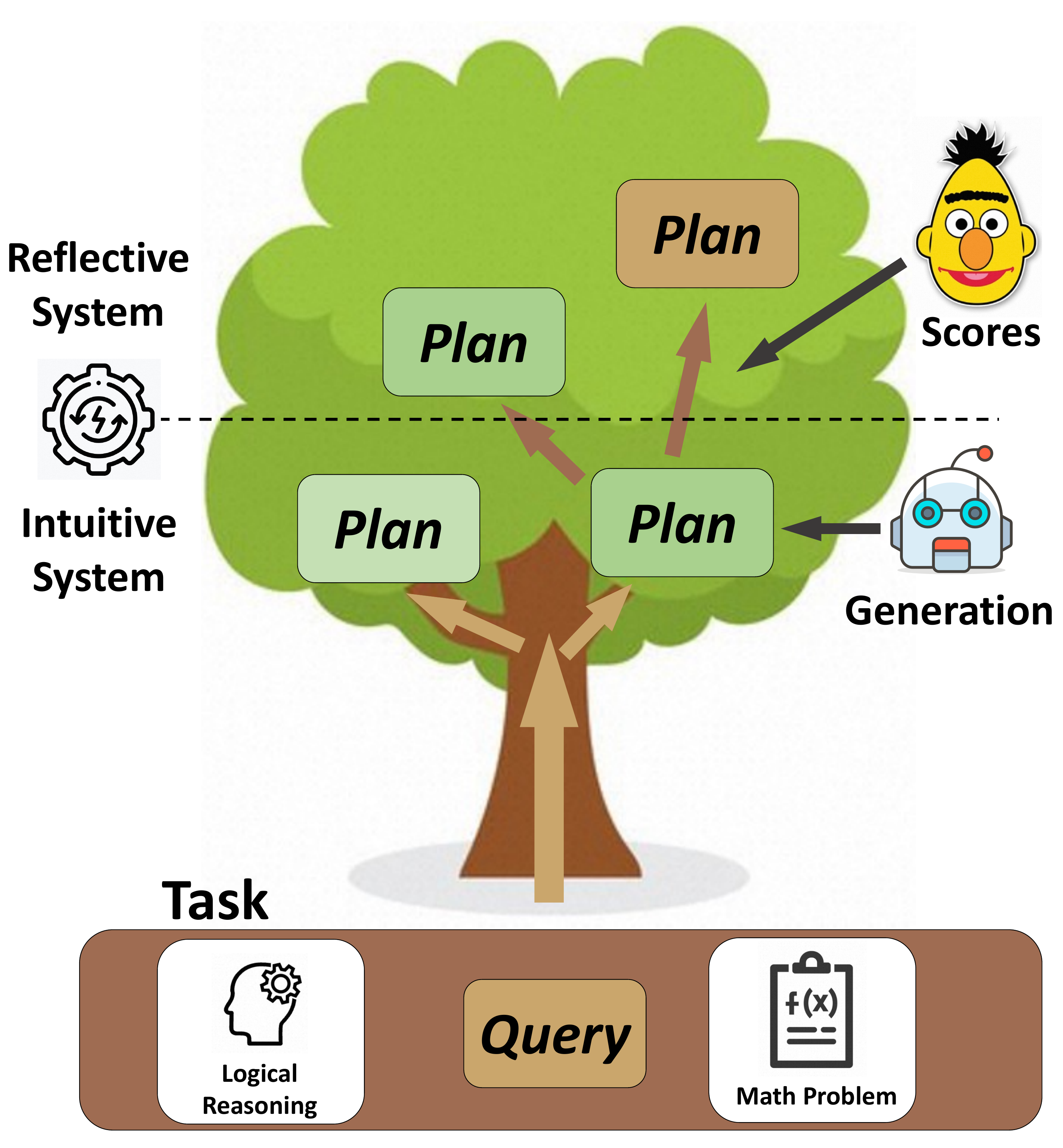}
\caption{A schematic illustration of the proposed framework named CogTree. An intuitive system is employed to generate candidate plans, while a reflective system verifies the plausibility of each plan to guide the next generation of the intuitive system. This iterative process is repeated to create the tree structure for reasoning.}
\label{motivation}
\end{figure}

We suggest that valuable insights into addressing these challenges can be derived from the cognitive processes of humans. 
In cognitive science, the~\emph{dual process theory}~\cite{evans1984heuristic,evans2003two,evans2008dualreview,sloman1996empirical} states that our brain initially employs an implicit, unconscious, and intuitive process known as the~\textbf{Intuitive System}, which retrieves relevant information. %following attention. 
This is followed by an explicit, conscious, and controllable reasoning process called the~\textbf{Reflective System}. 
The Intuitive System is capable of providing resources in response to requests, while the Reflective System facilitates a deeper exploration of relational information through sequential thinking in the working memory. Although slower, the Reflective System possesses a unique human rationality~\citep{baddeley2010working}. In complex reasoning tasks (including logical reasoning and mathematical reasoning tasks), these two systems coordinate with each other, engaging in iterative cycles of \textit{fast and slow} thinking~\cite{daniel2017thinking}.

In this paper, we propose the \textbf{Cognitive Tree} (CogTree) framework to address the aforementioned issues. Inspired by the dual process theory, our system consists of the Intuitive System and the Reflective System. 
In our implementation, the Intuitive System and the Reflective System are both generative models, albeit with distinct objectives. The Intuitive System employs in-context examples to dissect intricate problems into sub-problems and produce responses to the query. Conversely, the Reflective System evaluates the outcomes generated by the Intuitive System and chooses the most likely solution to provide guidance for the next generation step.
% The Intuitive System is modeled as a generative language model that employs contextual decomposition of examples to break down complex problems into sub-problems and generates answers to questions. In order to maintain conciseness in our implementation, we consistently utilize the same generative model for the Reflective System, producing different outcomes, which further comprehends the results from the Intuitive System, assesses their confidence, and guides the Intuitive System in the next step of decomposition. 
The aforementioned process is an iterative tree generation process, which continues until the original problem is decomposed into manageable sub-problems (corresponding to nodes on the tree) that can be easily solved.

Our main contributions are as follows:
\begin{itemize}

\item \textbf{Problem Decomposition Paradigm.} We propose a novel framework based on human cognition called Cognitive Tree (CogTree) for solving complex reasoning problems.

\item \textbf{Improved Validation Capabilities.}
By exposing the model to contrastive examples of correct decisions versus incorrect or ambiguous ones, we can improve the model's ability to make decisions (cognition ability). 
Additionally, apart from evaluating the model's judgment of intermediate results, we have also integrated the model's assessment of the overall correctness of the reasoning process.
% Alongside process supervision, we utilize a global supervision approach to enhance the model's long-term planning abilities. This allows the model to consider the broader context, anticipate future steps, and make strategic decisions in the problem-solving process.
% \textcolor{red}{
% The para. is not well aligned with intro (the two systems).}

\item \textbf{Efficient Framework.} The combination of the Intuitive System and the Reflective System can be applied to various reasoning tasks, i.e., both logical reasoning and mathematical reasoning. Notably, we are able to attain comparable reasoning performance to models with substantially larger parameter sizes, such as GPT-3.5 with 175B parameters, while utilizing relatively small language models (with 1.5B and 7B parameters).
This allows the trained small models to be deployed online for efficient inference.
% \textcolor{red}{
% need to emphasize that we can do reasoning with small models, both here and intro. scalable means that our work can be implemented on very large models,
% which is not suitable here. }

\end{itemize}

\begin{figure*}
\centering
\includegraphics[width=\textwidth]{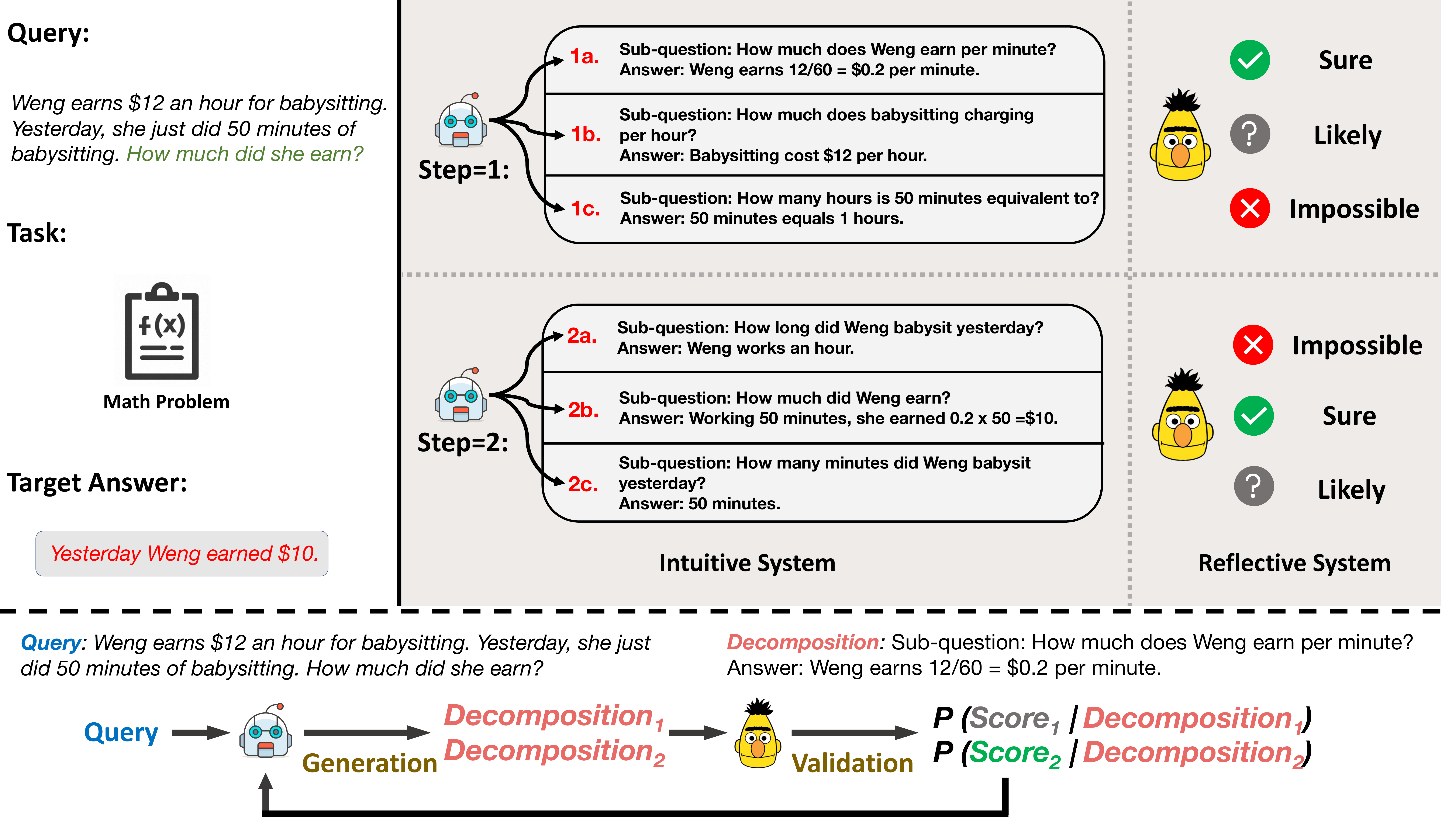}
\caption{An illustration of how the Intuitive System and the Reflective System work to incrementally produce a mathematical reasoning problem. At each step, the Intuitive System generates a set of decompositions based on the query. The Reflective System then scores the candidate decompositions and returns the top-ranked ones. The process terminates when the decomposition successfully matches the target answer.}
\label{motivation}
\end{figure*}

\section{Cognitive Tree (CogTree) Framework}

% The reasoning ability of humankind largely stems from the acquisition of relevant information from the surrounding environment and subconscious processing~\cite{think-sbs}.
% Our approach integrates the use of a tree structure to sequentially address reasoning problems, drawing inspiration from human problem-solving procedure, and follows a similar methodology to the planning processes examined by~\citealp{probsolve1959, probsolve1972}. 
The reasoning ability of humans primarily arises from acquiring pertinent information from the environment and subconscious processing~\cite{think-sbs}. In our approach, we incorporate a tree structure to systematically tackle reasoning problems, taking inspiration from human problem-solving procedures. This methodology aligns with the planning processes analyzed by~\citealp{probsolve1959, probsolve1972}.
Newell and his colleagues defined problem solving~\cite{probsolve1959} as the exploration of a combinatorial problem space, represented as a tree.
% \textcolor{red}{why tree for humans? no support here}

In our mathematical and logical reasoning setting, each node $n$ in the \textit{cognitive tree} $\mathcal{T}$ represents either a theory in a logical set or the solution to a sub-problem in a mathematical question. An edge $e$ of the tree corresponds to the evaluation of the current node's state $s$, which can be a confidence score or a classification result.
% \textcolor{red}{maybe we need to point out in intro that our approach can solve both math and logic problems.}
The problem decomposition module (i.e., the Intuitive System) receives the theory and the original problem as input, and generates the decomposition of the original problem.
Next, the newly generated nodes are used to expand the tree, providing the reasoning module (i.e., the Reflective System) with the information that needs to be verified.
% \textcolor{red}{maybe we need a figure to show the tree explicitly. another issue. we have discuss what the node is. how about edges?}

The discernment capacity of the Reflective System plays a pivotal role in enhancing the overall efficacy of the model~\cite{mathprompt}. In particular, we utilize the cross-checking technique to not only verify the precision of intermediate outcomes but also validate the accuracy of the entire reasoning process upon its completion. However, relying solely on individual decision judgments for training the model is inadequate~\cite{mathprompt}. To augment the model's ability to evaluate the state $s$, we propose the implementation of a comparative reinforcement approach. This approach entails introducing a new training objective, whereby the model is tasked with maximizing the disparity in vector space between representations of correct decisions and representations of incorrect or ambiguous decisions.
% By means of comparison, cognitive capabilities can be attained. The contrasting effect holds notable influence in the realm of judgment and decision-making. Individuals have a tendency to juxtapose the present state with past encounters or established benchmarks in order to gain a deeper comprehension and evaluation. Within Reflective System, assimilating lessons from contrasting instances of accurate and erroneous reasoning aids in assessing the viability of the current nodes $s$. Moreover, to ensure thorough validation of the entire chain of reasoning, Reflective System identifies any erroneous steps within the sequence, facilitates recuperation from those errors, and sustains the generative process.
% \textcolor{red}{too much cognitive stuff here. how about directly discuss the validation part? be concise. reviewers may not understand this part.}

In addition, efficiency pertains to the efficacy of the problem decomposition and verification process.
%Chain-of-Thought's effectiveness depends on the contextual examples provided, while Tree-of-Thought goes a step further by generating and verifying multiple answers. 
Generating lengthy texts using a model as massive as 175B entails substantial time and financial expenses. To tackle this issue, our method can be implemented by simply fine-tuning a comparatively smaller model (1.5B or 7B) exclusively for reasoning tasks. This enables us to deploy the model for these tasks with minimal costs.

\section{Implementation}

In this section, we describe the implementations of CogTree in detail.

%We state some basic notations.
%Denote $f_\theta$ as a decoder-based language model parameterized by $\theta$,
%and an input token sequence as $X=\{x_1,\cdots, x_i,\cdots, x_n\}$, where $n$ is the sequence length. We have $f_\theta(x) = \prod_{i=1}^{n} f_\theta(x_i | x_{1,...,i-1})$.

%We use $\mathcal{T}$ to denote a cognition tree where each node is a state $s$.
\subsection{Intuitive System}

The generative capability of the Intuitive System serves as the foundation for constructing the Cognitive Tree.
Thus, we choose decoder-only models (e.g., GPT2-XL~\cite{gpt2} or LLaMA-7B~\cite{LLaMA}) as the Intuitive System.

% To enhance the effectiveness of the Intuitive System, we adopt an in-context approach.
% Define the Query $\mathcal{Q}$ as the ultimate goal that needs to be proven for logical reasoning problems, or the question that needs to be answered for mathematical problems. Decomposition $\mathcal{D}$ in logical reasoning problems refers to further breaking down the goal, where reasoning through the obtained decomposition leads to the goal. In mathematical problems, it refers to one of the subproblems of the original problem, and solving this subproblem helps resolve the original problem. The Decomposition set $Z$ refers to the collection of decompositions for all examples in the training set.
%  $K$ examples (e.g. Query: $\mathcal{Q}$; Decomposition: $\mathcal{D}$) are recalled from the inference decomposition set $Z$ and used as the context for the model's input. 

\begin{figure}[tbp]
\centering
\includegraphics[width=1\columnwidth]{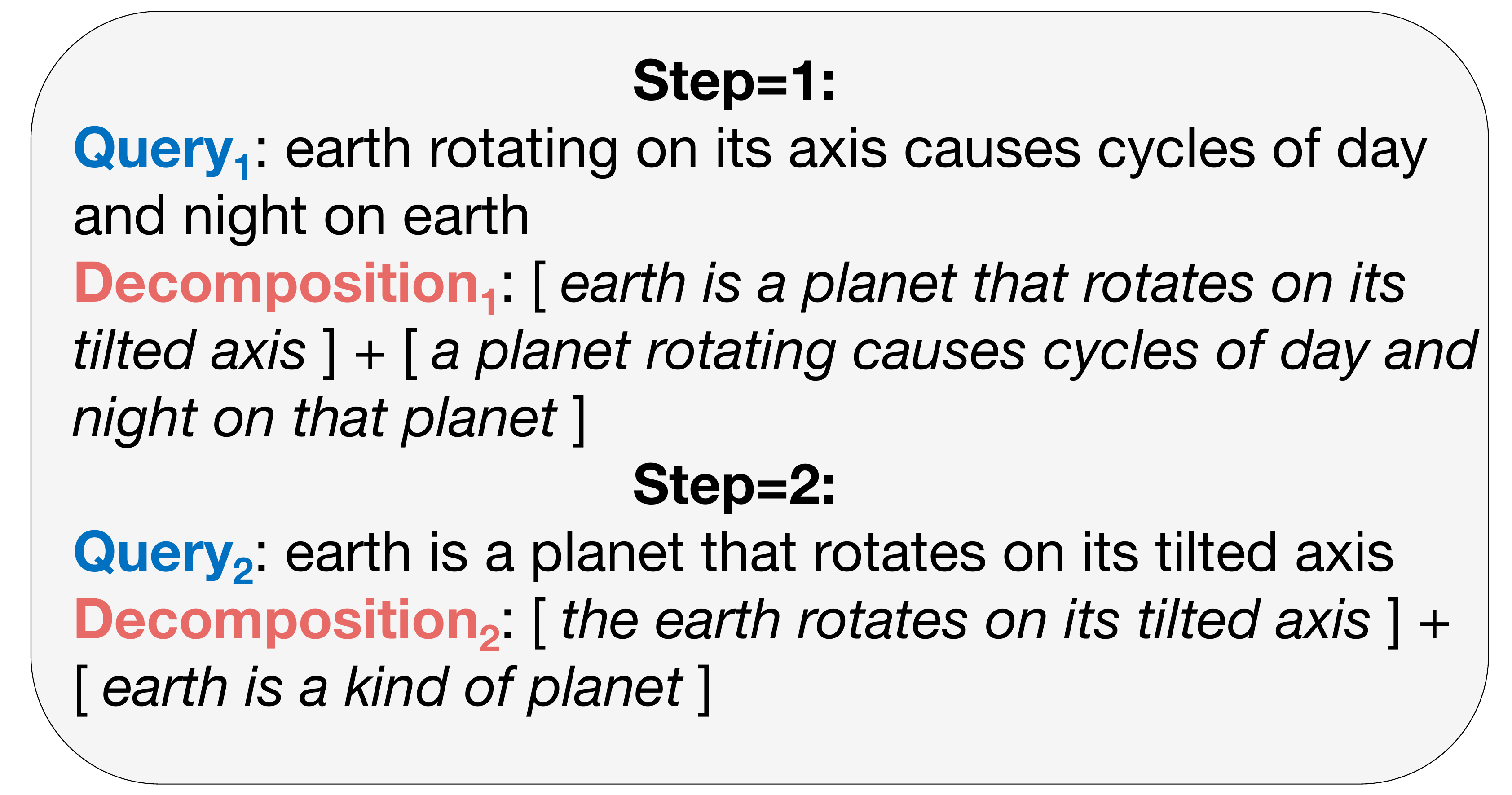}
\caption{Example of Query and Decomposition for logical reasoning.}
\label{fig:exampleEB}
\end{figure}

\begin{figure}[tbp]
\centering
\includegraphics[width=1\columnwidth]{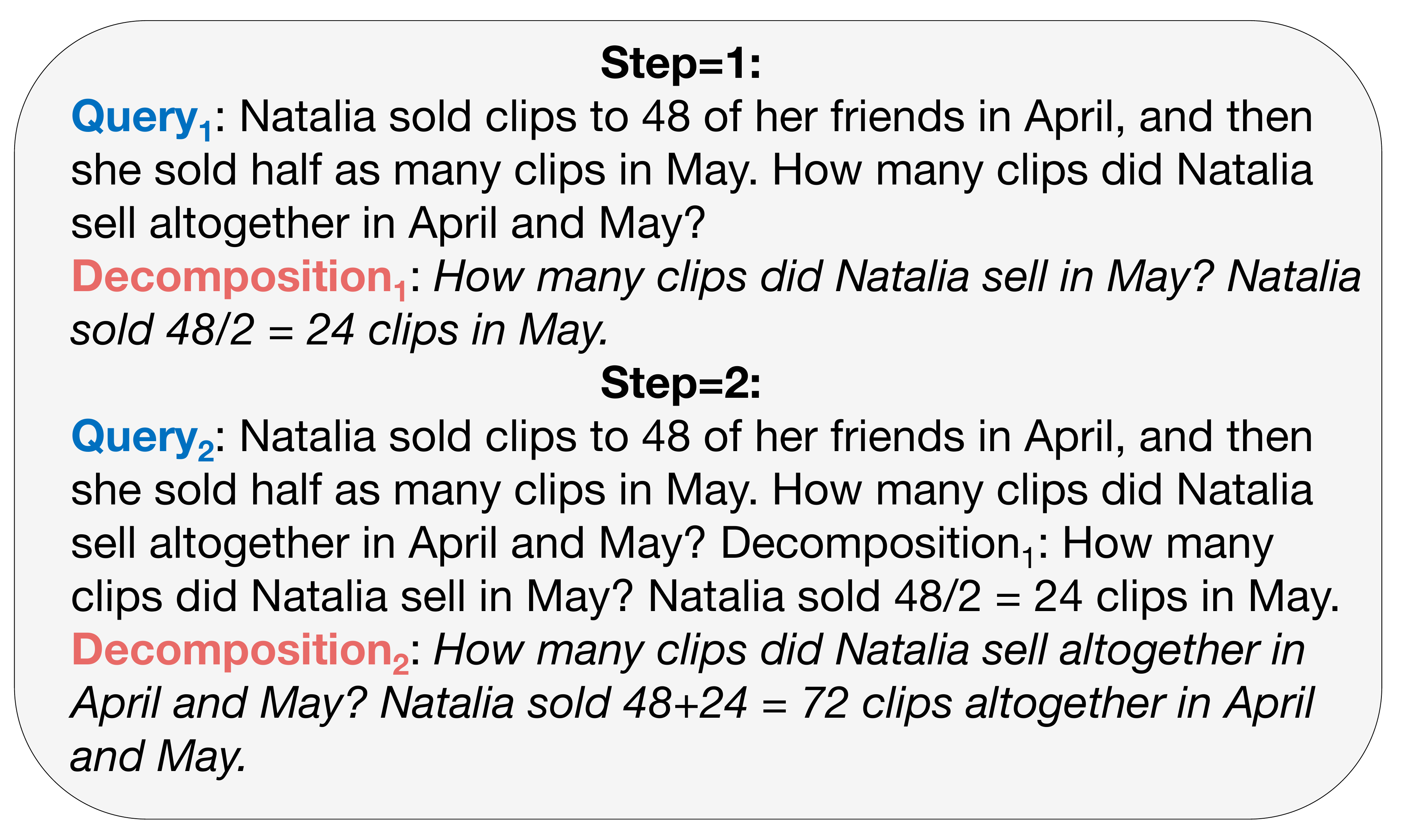}
\caption{Example of Query and Decomposition for mathematical problems.}
\label{fig:exampleGSM8K}
\end{figure}

 To enhance the effectiveness of the Intuitive System, we employ an in-context approach. Let us define the Query ($\mathcal{Q}$) as the ultimate goal for logical reasoning problems or the question to be answered in mathematical problems. In the case of logical reasoning problems, the Decomposition ($\mathcal{D}$) involves further breaking down the goal into smaller components, where reasoning through this decomposition enables the attainment of the goal. For mathematical problems, it refers to one of the sub-problems derived from the original problem, and solving this sub-problem contributes to resolving the original problem as a whole. The Decomposition set ($Z$) represents the collection of decompositions for all examples in the training set. In our approach, we retrieve $K$ examples (e.g., Query: $\mathcal{Q}$; Decomposition: $\mathcal{D}$) from the inference decomposition set ($Z$), which are then utilized as the context for the model's input.
Detailed examples are shown in Figure~\ref{fig:exampleEB} and Figure~\ref{fig:exampleGSM8K}.
% \begin{equation*}
% \begin{small}
% \label{input}
%     \underbrace{ Example[1] \ | \ Example[2] \ | \ Question}_{Context} \underbrace{\ Decomposition}_{To\ be\ generated}
% \end{small}
% \end{equation*}

% \textcolor{red}{(lack of def, such as meta-question, sub-question, decomposition set)}

Then, the output can be generated as $y \sim f_\theta(y | x, z_{1 \cdots K})$. Here, $z$ represents the $K$ examples\footnote{In the experiment, we set $K=5$.} recalled from the decompositions set $Z$, where $Z=\{z_1,\cdots, z_L\}$. 
In practice, we use the Intuitive System to obtain the representation of the current query (final transformer block's activation) and calculate the cosine similarity with the representations of other queries in set $Z$. We then retrieve the $K$ most similar queries from the set. Moreover, $[y] \sim f_\theta(y | x, z_{1 \cdots K})$ is sampled as a continuous language sequence.

\subsection{Reflective System}

% Reflective System differs from the quick intuition generation of Intuitive System, as its role is to verify the decompositions and answers provided by Intuitive System for their acceptability. The overall effectiveness greatly depends on the accuracy of Reflective System's judgments. In practice, we adopt two verification methods: verification of intermediate processes and verification of the entire reasoning chain. Based on the current state $s$, we can utilize Reflective System (the same model architecture as Intuitive System) to generate a value $v$ that validates the current state: $V(f_\theta,s) \sim f_\theta(v | s)$. Based on the whole reasoning chain $S=\{s_1,\cdots, s_i,\cdots, s_n\}$, we can utilize Reflective System to generate a overall assessment $o$: $O(f_\theta,S) \sim f_\theta(o | S)$.

% To generate the value $v$ and $o$, we offer two approaches: For logical reasoning problems, the model generates a confidence value (Integer values between 0 and 10) to indicate its level of certainty. For more complex mathematical problems, the answers generated by Intuitive System can sometimes be misleading and cannot be accurately judged at this stage. Therefore, we employ a prompt-based approach, treating it as a classification problem for the model to output \textit{positive}, \textit{negative}, or \textit{neutral}. The \textit{neutral} response signifies that the generated answer is plausible and requires further verification.

The Reflective System differs from the Intuitive System in terms of its approach of generating insights. While the Intuitive System relies on quick intuition, the Reflective System's role is to evaluate the decompositions to determine their acceptability. In practice, we employ two methods to verify the results: the verification of intermediate processes and the verification of the entire reasoning chain.

Given the current state $s$ (Query: $\mathcal{Q}$ with Decomposition: $\mathcal{D}$), we utilize the Reflective System, which shares the same model architecture as the Intuitive System, to generate a score $v$ that validates the current state. 
This is represented by $V(f_\theta,s) \sim f_\theta(v | s)$. Additionally, based on the complete reasoning chain $S=\{s_1,\cdots, s_i,\cdots, s_n\}$, we employ the Reflective System to produce an overall score $o$, which can be expressed as $O(f_\theta,S) \sim f_\theta(o | S)$.

% To generate the scores $v$ and $o$, we employ two approaches. 
% i) Regression: the model generates a confidence value ranging from 0 to 1 to indicate its certainty level. 
% The inputs are passed through our model to obtain the final transformer block’s activation $h_n^l$ , which is then fed into an added linear output layer with parameters $W_v$ to predict $v$, which is formulated as:
% \begin{equation}
% P(v | x_0, \cdots, x_n )=softmax(h_n^l W_v)
% \end{equation}
To generate the scores $v$ and $o$, we utilize the model to produce a classification result. 
Since the answers generated by the Intuitive System can sometimes be misleading and cannot be accurately assessed at this stage, we adopt a prompt-based approach and treat it as a classification problem, where the model outputs one of three categories: \textit{sure}, \textit{impossible}, or \textit{likely}. The \textit{likely} response signifies that the generated answer is plausible but requires further verification.
% \footnote{In the implementation, we adopt the classification approach to distinguish the results generated by the Intuitive System for both logical reasoning and mathematical tasks.}

\begin{table*}
  \centering
  \begin{small}
  \begin{tabularx}{\textwidth}{l|XX}
    \toprule
    \textbf{Dataset} &  \textbf{Input} & \textbf{Output} \\
    \midrule
    \textbf{EB} & A hypothesis that needs to be proven and a set of theory. {\textcolor{blue}{(Hypo: Phobos is a kind of moon. Theory: [Mars is a kind of planet; moons orbit planets; Phobos orbits Mars.])}}  & Yes or No to indicate whether or not the hypothesis can be proven based on the theory. {\textcolor{blue}{(Yes)}} \\
    \midrule
    \textbf{GSM8K} & A math word problem {\textcolor{blue}{(Natalia sold clips to 48 of her friends in April, and then she sold half as many clips in May. How many clips did Natalia sell altogether in April and May?)}} & A number denoting the solution to the mathematical problem. {\textcolor{blue}{(72)}}\\
    \bottomrule
  \end{tabularx}
  \end{small}
\caption{Task overview. Examples of input and output are printed in blue.}
\label{tab:task}
\end{table*}

\begin{table}[t]
\begin{center}
\begin{small}
\begin{tabular}{lcccc}
\toprule
\bf Dataset & \bf Split & \bf \# Samples & \bf Max Steps & \bf Avg Steps \\
\midrule
EB & Train & 1313 & 17 & 3.2 \\
& Dev & 187 & 15 & 3.2 \\
& Test   & 340 & 11 & 3.3 \\
\midrule
GSM8K & Train & 6726 & 9 & 3.6 \\
& Dev & 747 & 8 & 3.5 \\
& Test   & 1319 & 11 & 3.7 \\
\bottomrule
\end{tabular}
\end{small}
\caption{Data statistics of EB and GSM8K.}
\label{tab:stats}
\end{center}
\end{table}

\subsection{Training}
\noindent\textbf{Intuitive System.} Supervised Fine-tuning (SFT) has demonstrated its effectiveness in aligning with human intentions~\cite{instructGPT}. In our approach, the Intuitive System is designed to decompose the queries (i.e., complex problems) into sub-problems by leveraging in-context examples. Since we employ generative models as our Intuitive System, the loss calculation is only necessary for the generative text (without the given context) during auto-regressive computation. Given a sample of tokens with a length of $N$ denoted as $X$, where $X=\{x_1,\cdots, x_i,\cdots, x_n\}$. Furthermore, we define the sequence length of in-context examples as $M$. 
%The token length of Z is denoted as M. 
We use a standard language modeling objective to maximize the following likelihood function:
\begin{equation}
    \mathcal{L}_{\mathcal{IS}}=\sum_{i>M}^N log \ P(x_i | x_1, \cdots, x_{i-1}; \theta)
\end{equation}

\noindent\textbf{Reflective System.} 
The acquisition of the Reflective System can be achieved through the same training approach as the Intuitive System, which involves utilizing positive and negative samples to obtain classification results from the model. Since the Reflective System is primarily focused on generating judgments for a given state $s$, the loss function can be defined as follows:
\begin{equation}
\mathcal{L}_{\mathcal{RS}} = \log P(v | s; \theta)
\end{equation}

However, the effectiveness of this training method  is found to be unsatisfactory. In cognitive theory, human decision-making behavior arises from the comparative analysis of various options~\cite{CognitiveDissonance}. Drawing inspiration from the cognitive theory, we adopt a contrastive learning approach to enhance the model's ability to distinguish between different states. The fundamental concept of contrastive learning is to learn representations of positive and negative samples by maximizing their distance in the sample space~\cite{simclr}. Consequently, the selection of negative samples plays a critical role in determining the effectiveness of contrastive learning.

For logic reasoning datasets, one approach to generate more challenging negative examples is to replace one of the theories in decomposition with another theory from the current theory set. This negative example is more challenging for the model to distinguish because the theories within the same theory set are more similar.

For mathematical problems, since our experimental dataset GSM8K~\cite{gsm8k} only provides the correct answer itself, it does not offer incorrect solutions. We use the dataset PRM800K~\cite{prm800k} to enhance the learning process, where there are ambiguous responses 
 $S'=\{s'_1,\cdots, s'_i,\cdots, s'_n\}$ (seemingly correct but actually incorrect). The judgment generated by Reflective System is $V'=\{v'_1,\cdots, v'_i,\cdots, v'_n\}$. By maximizing the distance between $v'$ and the correct answers $v$, we can enhance the learning process.
Let $g(v',\cdot)$\footnote{In the experiment, we use cosine similarity as $g(v',\cdot)$} be a matching function between negative sample $v'$ and the positive sample $v$. The loss function for contrastive learning can be expressed as follows:
\begin{equation}
\mathcal{L}_{\mathcal{CL}} = \frac{\exp(f(v,y))}{\exp(f(v,y)) + \sum_{v'\sim V'}\exp (f(v,v'))}
\end{equation}

Hence, the total loss function for the Reflective System is given by:

\begin{equation}
\label{eq:loss}
\mathcal{L}_{\mathrm{total}}=\lambda\cdot\mathcal{L}_{\mathcal{RS}}+(1-\lambda)\cdot\mathcal{L}_{\mathcal{CL}}
\end{equation}
Here, $\lambda$ is the hyper-parameter. We conduct experiments on $\lambda$ and choose $\lambda=0.5$ as the best setting.\footnote{For specific details, please refer to the Section~\ref{exp:hyper}.}

\section{Experiments}
We perform an extensive evaluation of our method utilizing two well-established benchmark datasets: the Entailment Bank (EB)~\cite{entailment2021} and GSM8K~\cite{gsm8k}. EB consists of human-annotated tuples containing information about theories, provable goals, and corresponding reasoning paths. GSM8K, on the other hand, presents a challenging arithmetic reasoning task that language models frequently find difficult to tackle~\cite{math2021, mathword}.
Examples of these two datasets are in Table~\ref{tab:task}. The dataset statistics are shown in Table~\ref{tab:stats}.

\begin{table*}[th!]
  \centering
  \begin{small}
  \begin{tabular}{lc|cc|cc}
      \toprule
        & & \multicolumn{2}{c}{\textbf{EB}}& \multicolumn{2}{c}{\textbf{GSM8K}} \\
       \cmidrule(lr){3-6}
        \textbf{Model}
        & \textbf{\#Params.} 
        &   \textbf{Accuracy (\%)} & $\mathbf{\Delta}$ (\%)
        &   \textbf{Accuracy (\%)} & $\mathbf{\Delta}$ (\%)\\
      \midrule
      \textbf{Comparative Systems}\\
      \midrule
        %%%
        GPT-3.5 (code-davinci-002) & 175B 
        &  80.76 & - &  16.17 & -\\ 
        %%%
        %%%
        \;\;\;\; + Standard prompt  & 175B 
        &  84.23 & +3.47&  17.03 & +0.86 \\ 
        %%%
        %%%
        \;\;\;\; + Chain-of-thought prompt & 175B 
        &  92.45 & +11.69& 60.27& +44.10\\ 
        %%%
        \;\;\;\; + Tree-of-thought prompt & 175B 
        &  93.31 & +12.55& \textbf{61.39}& +45.22 \\ 
      \midrule
      \textbf{Our Models (CogTree)}\\
      \midrule      GPT2-XL (Intuitive System only)
      & 1.5B 
      &  82.37 & - & 23.53& -\\

      \;\;\;\; + GPT2-XL (as Reflective System) 
      &  1.5B  
      &  92.63  & +10.26 & 35.84& +12.31\\
      
      \;\;\;\; + LLaMA (as Reflective System)
       &  7B  
      & 93.16  &+10.79  &34.68 & +11.15 \\
      \midrule

      LLaMA (Intuitive System only)
      & 7B 
      &  86.14 & - & 43.52& -\\ 

    \;\;\;\; + GPT2-XL (as Reflective System) 
      &  1.5B  
      &  93.25  &+ 7.11 &47.80 & +4.28\\
      
    \;\;\;\; + LLaMA (as Reflective System)
      & 7B  
      &  \textbf{94.25} & +8.11& 61.28& +17.76  \\ 
      \bottomrule
  \end{tabular}
  \end{small}
  \caption{
  Overall test set performance in terms of accuracy and relative improvement.
  }
  \label{tab:exp}
\end{table*}

\subsection{Experimental Setup}
In our experiments, we use GPT2-XL~\cite{gpt2} and LLaMA-7B~\cite{llama} from the Huggingface transformers~\cite{huggingface} library as the underlying models for the Intuitive System and the Reflective System.

During training, we use the Adam optimizer with $\beta_1=0.9, \beta_2=0.999, \epsilon=1e-8$. We use the  learning rate $\gamma=1e-4$ for both Systems. We use a batch size of 4 and set a large epoch number (i.e., 100) and use the validation set to do early stopping. In practice, the best epoch is often within 50.
During the inference stage, in each step, we use the Intuitive System to generate the top\_beam=3 answers and then let the Reflective System select the most probable answer to continue generating for the next step.
We add an ``end'' marker. The inference process stops when the Intuitive System generates the ``end'' marker or when the maximum number of inferences, which is 20, is reached.
For each Query, we perform 5 complete reasoning process.\footnote{Detailed examples are shown in Figure~\ref{sucEB} and Figure~\ref{sucGSM8K}.}

Following~\citealp{least2most}, we use code-davinci-002\footnote{\url{https://openai.com/}} (code-davinci-002 is constructed on the foundation of the GPT-3.5 architecture) for comparative experiment due to its strong reasoning abilitiy and employ various prompt strategies to conduct our experiments on GPT2-XL and LLaMA-7B. For each example, we sample Standard prompting (detailed cases to be described in Section \ref{baseline}) and Chain-of-thought (CoT) prompting for 100 times for average performance. For Tree-of-thought (ToT), at each step we generate 5 candidate answers and sample values 3 times for each example.

% \noindent\textbf{Standard prompting.}
% Standard prompting, popularized by~\citealp{few-shotlearner}, involves providing a language model with contextual input-output examples, thereby enabling it to generate predictions for test-time examples. This approach allows the model to directly produce the answer.

% \noindent\textbf{Chain-of-thought prompting.}
% The concept of Chain-of-thought~\cite{cot} was proposed as a solution to address scenarios where the mapping from input to output is not straightforward. This is particularly applicable when input represents a mathematical problem and output corresponds to the ultimate numerical solution. The fundamental approach involves the introduction of a sequence of thought chains, which establish a connection between the input and output. Each of these thought chains consists of a coherent sequence of language and serves as a meaningful intermediary step in the process of problem-solving.

% \noindent\textbf{Tree-of-thought prompting.}
% Tree-of-thought~\cite{tot} is a novel tree-based reasoning approach that expands the model's inference process beyond linear reasoning. It converts the inference process into a tree-based search. 
% %Inspired by the Chain-of-thought (COT) concept, 
% It enables the model to generate multiple reasoning paths during result generation, then selects the most likely path to proceed with the generation process, thereby enhancing the model's overall performance.

\subsection{Results on Entailment Bank}
On EB, followed by~\citealp{ExplicitPlanning2023}, we assess the capabilities of systems in distinguishing between provable and non-provable goals. To accomplish this, we assign a non-provable goal to each development and testing theories by selecting it from other (theory, goal, reasoning path) samples.
The selection is adversarial: We input all the goals in the set into our pre-trained model separately, obtaining the last output of the last transformer block as the representation of each goal. We proceed by computing the cosine similarity between all non-provable goals and the provable goal. Based on this computation, we identify the hard negative example with the highest similarity.
For a given theory $T$ and a query $Q$, we allow the system to generate a reasoning path $S$
%aimed at proving the goal 
and obtain the proof value $o=f_\theta(o | S)$ for that path. Given the choices ``Sure/Likely/Impossible'', we say ``$Q$ is provable'' when the value $o$ is ``sure'' and not provable otherwise.

\noindent\textbf{Baselines.}
\label{baseline}
\textbf{Vanilla}: The raw input to the model is as follows: Query: $\mathcal{Q}$; Theory Set: $\mathcal{T}$, followed by a question: Based on the theory, can the goal be approved? Please answer with yes or no.
\textbf{Standard prompt}: We use an input-output (IO) prompt with 5 in-context examples (e.g. Query: $\mathcal{Q}$; Theory Set: $\mathcal{T}$; Answer: $\mathcal{A}$.).
\textbf{Chain-of-though prompt}: For chain-of- thought (CoT) prompting, we augment each input-output pair using examples with complete reasoning chain (e.g. Query: $\mathcal{Q}$; Theory Set: $\mathcal{T}$; Reasoning Chain: $\mathcal{S}$; Answer: $\mathcal{A}$.).
\textbf{Tree-of-thought prompt}: In each step, we provide the model with the theory set and ask the model to select two theories from the set to generate a new inference, which is added to the theory set while removing the two selected theories. Each step generates five candidates, and the model generates subsequent inferences based on the highest likelihood until the set ultimately contains only one inference, which is compared with the goal to determine consistency (e.g., theory$_{i}$ + theory$_{j}$ -> inference (theory set without theory$_{i \& j}$)).

% \begin{figure}[tbp]
% \vspace{-.5em}
% \centering
% \includegraphics[width=0.8\columnwidth]{exp/error_bar_eb.pdf}
% \vspace{-.5em}
% \caption{Samples failed at each step on EB.}
% \vspace{-1em}
% \label{fig:freq}
% \end{figure}

\begin{figure}[tbp]
\centering
\begin{tabular}{ll}
\begin{minipage}[t]{0.48\linewidth}
    \includegraphics[width = 1\linewidth]{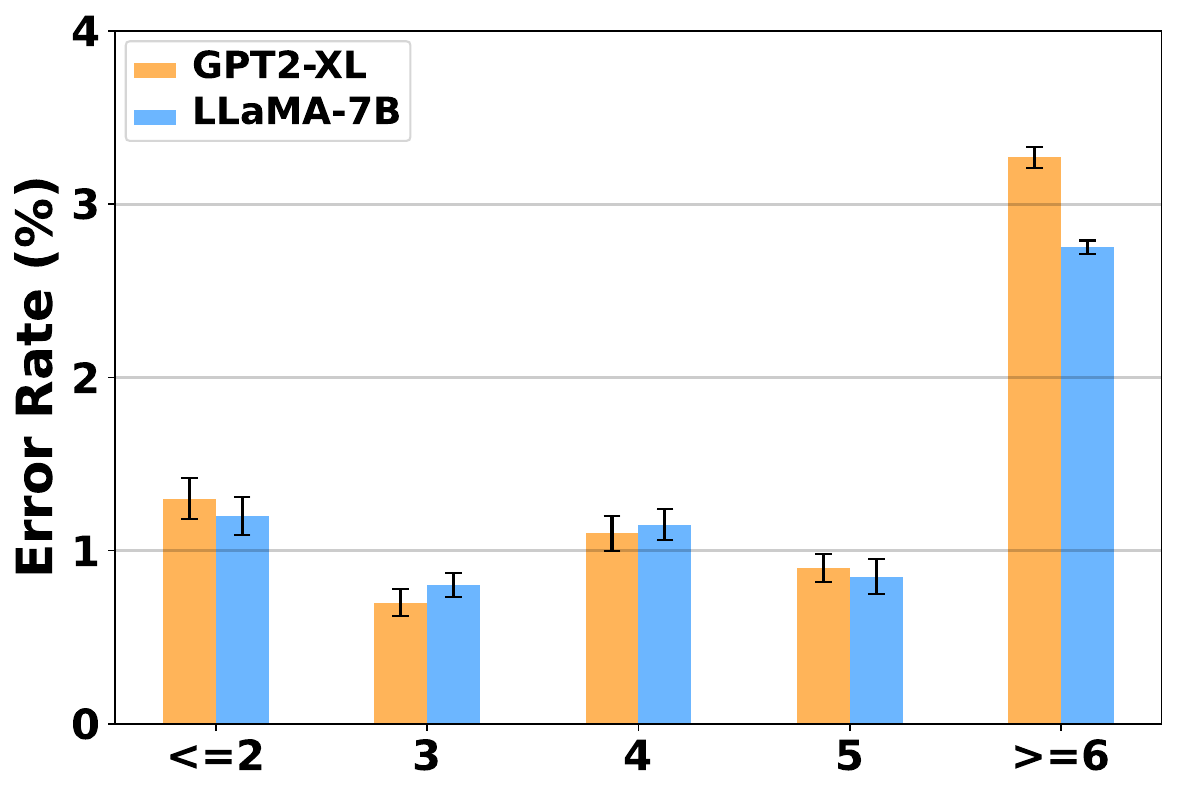}
\end{minipage}
\begin{minipage}[t]{0.48\linewidth}
    \includegraphics[width = 1\linewidth]{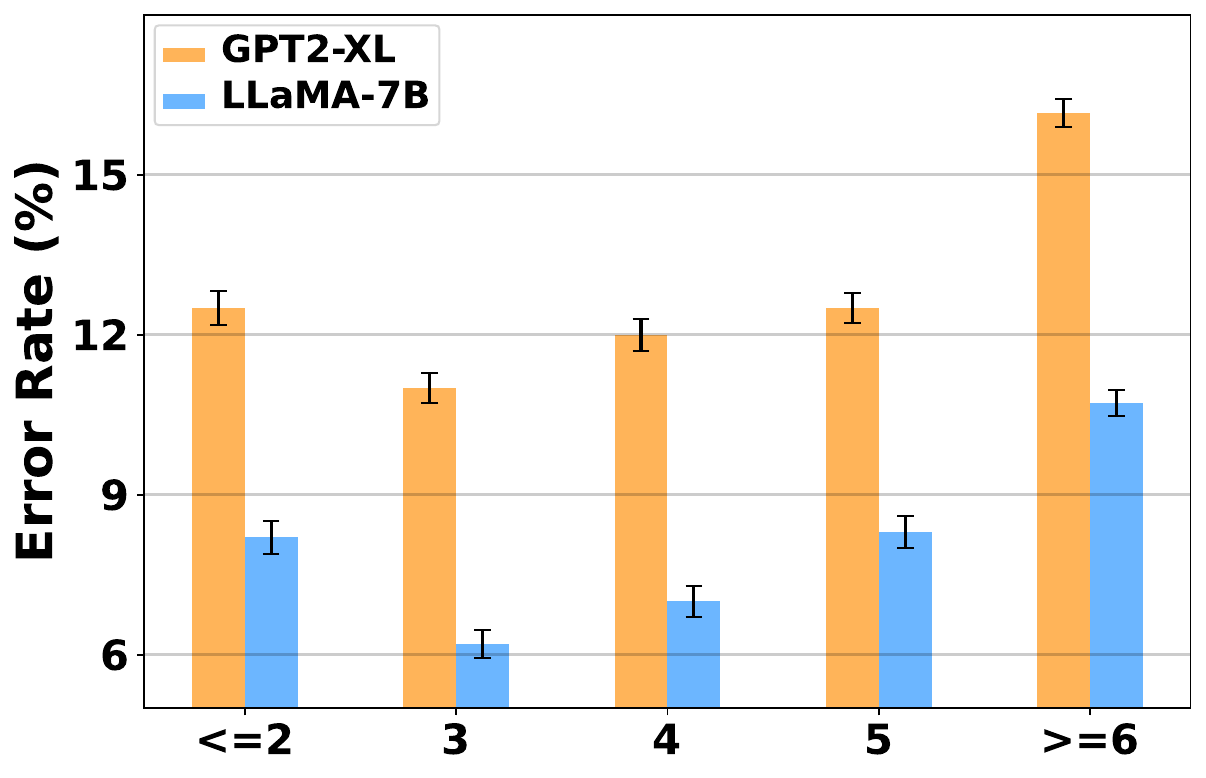}
\end{minipage}
\end{tabular}
\caption{Samples failed at each step on EB (left) and GSM8K (right).}
\label{fig:error}
\end{figure}

\noindent\textbf{Main results.}
Table~\ref{tab:exp} presents the classification accuracy on EB. The results demonstrate that GPT2-XL (1.5B), trained exclusively on in-context examples, outperforms GPT-3.5 (175B) despite having fewer than 1\% of the model's parameters. By incorporating CoT and ToT augmentation methods, the accuracy of GPT-3.5 is substantially enhanced, reaching 92-93\%. Furthermore, when our approach is combined with the Reflective System for result verification, even higher performance is achieved (94\% by LLaMA-7B), surpassing the prompt augmentation method employed by GPT-3.5.

\noindent\textbf{Error Analysis.}
Although our model has achieved satisfactory results, it still exhibits deficiencies in certain cases. We have conducted an examination of instances where the model's inference has failed (Detailed examples are shown in Figure~\ref{fig:error}). Specifically, we observed a decline in the model's accuracy when the length of the inference chain exceeds 10 steps. Our model struggles to determine the appropriate decomposition of the goal in such cases. This limitation may arise from the excessively long and divergent nature of the chain required to reach the goal, which surpasses the model's current capacity for abstraction in this regard. It is worth noting that even for humans, this task can be complex, often requiring multiple attempts to arrive at the final inference chain.

\subsection{Results on GSM8K}
%On the GSM8K dataset, we expect our model to output the answer (a number) to a given question.

On GSM8K, we employ in-context examples
%(e.g., problem: p; sub-problem: p) 
to enable the model to generate sub-questions for a given problem. In the implementation, we conduct five sampling iterations at each decomposition step and choose the one with the highest probability for generating the next step. Once all the sub-questions have been generated, we proceed to have the model answer each sub-question sequentially, ultimately deriving the final answer.

\noindent\textbf{Baselines.}
\textbf{Vanilla}: Directly input the question and let the model to answer (e.g. Query: $\mathcal{Q}$).
\textbf{Standard prompt}: We use an input-output (IO) prompt with 5 in-context examples (e.g. Query: $\mathcal{Q}$; Answer: $\mathcal{A}$).
\textbf{Chain-of-though prompt}: For CoT, we use the step-by-step solution of mathematical problems as enhanced input-output pair (e.g. Query: $\mathcal{Q}$; Reasoning Chain: $\mathcal{S}$; Answer: $\mathcal{A}$).
\textbf{Tree-of-though prompt}: We adopt the methodology described in ToT~\cite{tot}. Our implementation involves the incremental generation of the answer, with the model producing one step at a time. We sample the results five times for each step and employ the model to assess the generated outcomes until the final answer is obtained\footnote{It is worth noting that our tree construction method differs significantly from ToT~\cite{tot}. ToT employs a bottom-up approach, whereas we utilize explicit questioning to break down the problem for the model and address it in a top-down fashion, step by step.}. 

% \begin{table}[t]
% %\vspace{-1em}
% \centering
% \begin{small}
% \begin{tabular}{c|cccc}
% \toprule
% \multirow{2}*{Models} & \multicolumn{2}{c}{Entailment Bank} & \multicolumn{2}{c}{GSM8K}  	 \\
% \cmidrule(r){2-3} \cmidrule(r){4-5}
% ~ &\textbf{Acc.} & $\mathbf{\Delta}$
%         &   \textbf{Acc.} & $\mathbf{\Delta}$\\
% \midrule
% GPT2-XL (IS) & 82.37   &-  &23.53 &-  \\
% LLaMA-7B (RS) & 93.16  &+10.79  &34.68 & +11.15  \\ \midrule
% LLaMA-7B (IS) & 86.14 &-  &43.52 & -  \\
% GPT2-XL (RS) & 93.25  &+ 7.11 &47.80 & +4.28 \\
% \bottomrule
% \end{tabular}
% \end{small}
% \vspace{-.5em} 
% \caption{Backbone Modifications for IS (Intuitive System) and RS (Reflective System).}
% \vspace{-1em} 
% \label{exp:ISRS}
% \end{table}

\noindent\textbf{Main Results.}
Experimental results are shown in Table~\ref{tab:exp}. It is evident that the direct question-answering accuracy of GPT-3.5 is merely 16\%. The traditional input-output approach did not improve its effectiveness. In contrast, CoT and ToT demonstrate a significant enhancement in solving mathematical problems, with an improvement of approximately 44\%. This indicates the crucial role of providing the model with example reasoning chains for tasks involving multi-step inference. Our SFT-improved GPT2 achieves only 23.5\% accuracy, which may be attributed to the scale low~\cite{scalelaw} caused by model parameters. The performance of our LLaMA-7B and GPT-3.5 models is nearly indistinguishable, suggesting that fine-tuning small models using our provided method can approach the performance of larger models. Notably, the inclusion of the Reflective System on GSM8K leads to a greater overall improvement compared to EB, indicating that using the Reflective System can yield better results, particularly in more complex problems.

\begin{table}
  \center\small
  \begin{tabular}{l cc}
  \toprule
  \textbf{Model}     & \textbf{EB} & \textbf{GSM8K}  \\ \midrule
  \textbf{GPT-XL} \\ \midrule
  
  \;\;\;\;+Specialize &	81.54\%&	21.43\%  \\ 
  \;\;\;\;+DecomP&	83.27\%&	24.35\% \\
  \;\;\;\;+Self-Ask&	82.69\%&	25.46\% \\
  \;\;\;\;+CogTree&	\textbf{93.16\%}&	\textbf{34.68\%}\\
  \midrule
  \textbf{LLaMA-7B}  \\ \midrule
  
 \;\;\;\;+Specialize	&84.32\%&	34.21\%\\ 
 \;\;\;\;+DecomP&	82.43\%&	39.75\%\\
 \;\;\;\;+Self-Ask	&83.98\%	&38.35\%\\ 
 \;\;\;\;+CogTree	& \textbf{94.25\%}	&\textbf{61.28\%}\\
 \bottomrule
  \end{tabular}
  \caption{Performance metrics for different methods with EB and GSM8K.}
  \label{tab:sota}
\end{table}

\begin{table}
\centering
\begin{small}
\begin{tabular}{l cc}
\toprule
Model & EB & GSM8K \\ \midrule
GPT-XL & \bf 92.63\% & \bf 35.84\% \\ 
\quad \textbf{-} decomposition  & 62.38\%  &  	18.94\%  \\ \midrule
LLaMA-7B & \bf 93.25\% & \bf 61.28\% \\ 
\quad \textbf{-} decomposition  & 69.79\%  &  	27.16\%
\\
\bottomrule
\end{tabular}
\end{small}
\caption{Ablation study in terms of F1. }
\label{ablation_study}
\end{table}

\noindent\textbf{Error Analysis.}
We have found that the examples where the model fails to answer can be divided into two categories. The first category is the failure in decomposing the question into subproblems (Detailed examples are shown in Figure~\ref{fig:error}), which is consistent with the findings of \citet{least2most}. Such failures can be resolved through manual decomposition. The second category of failure is when the model provides inaccurate answers to the subproblems. These failures are frequently observed in GPT2-XL and are the main reason for the unsatisfactory performance of GPT2. One possible reason for these failures is the inadequacy of the model's parameter size, which hinders its ability to acquire fundamental mathematical capabilities.

\subsection{Backbone Modifications}
We conduct experiments employing different backbones for the Intuitive System and the Reflective System. The results presented in Table~\ref{tab:exp} demonstrate that when GPT2-XL is utilized as the Intuitive System and   LLaMA-7B as the Reflective System, there is an observed improvement in overall performance on the EB dataset, compared to using GPT2-XL as the Reflective System. However, when a more powerful model is employed as the Reflective System on the GSM8K dataset, there is no noticeable performance enhancement. This finding suggests that the Intuitive System restricts the system's performance on the GSM8K dataset. Conversely, when LLaMA-7B serves as the Intuitive System and GPT2-XL as the Reflective System, the performance improvement on the GSM8K dataset, in comparison to using LLaMA-7B as the Reflective System, is not substantial. This indicates that in this particular case, the Reflective System limits the overall system's performance.

\subsection{Compared with Other Finetune SOTAs}
In pursuit of fairness, we conducted a comparison of state-of-the-art methods that had been fine-tuned on identical datasets, in contrast to our prior evaluation of methods relying on GPT-3.5, which is parameter-free.

\noindent\textbf{DecomP}~\citep{DecomP} solves complex tasks by decomposing them (via prompting) into simpler sub-tasks.
\textbf{Specialize}~\citep{Specialize} proposes small model specialization to enhance performance by focusing model capacity on a specific target task.
\textbf{Self-Ask}~\citep{Self-Ask} explicitly asks the model itself (and then answers) follow-up questions before
answering the initial question.

It can be observed in Table~\ref{tab:sota}, when compared with the finetune-based methods, Cogtree still achieves state-of-the-art results. This is attributed to not only decomposing the problem but also incorporating result validation in subsequent steps.

\subsection{Ablation Study}
We conducted further experiments to demonstrate the effectiveness of decomposing step by step. "w/ decomposition" indicates that we used the method from our paper to decompose and sequentially answer the original problem. On the other hand, "w/o decomposition" means that we did not use the intermediate problem decomposition and directly answered the original problem, relying on System 1 to generate the answer.

As we can see in the Table~\ref{ablation_study}, the accuracy of directly answering the original problem is low, especially when the original problem is complex. This is also in line with human cognition. When solving math problems, we also solve intermediate problems first and then obtain the final answer, which improves our accuracy.

\subsection{Hyper-parameter Analysis}
\label{exp:hyper}
\begin{figure}[tbp]
\centering
\begin{tabular}{ll}
\begin{minipage}[t]{0.48\linewidth}
    \includegraphics[width = 1\linewidth]{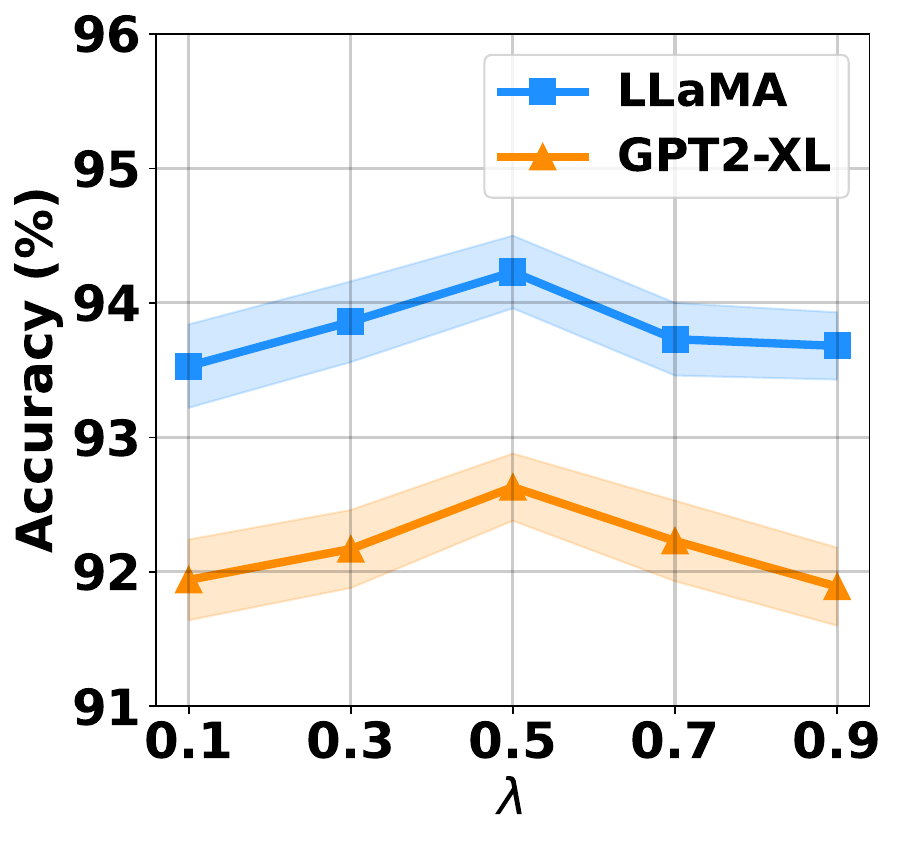}
\end{minipage}
\begin{minipage}[t]{0.48\linewidth}
    \includegraphics[width = 1\linewidth]{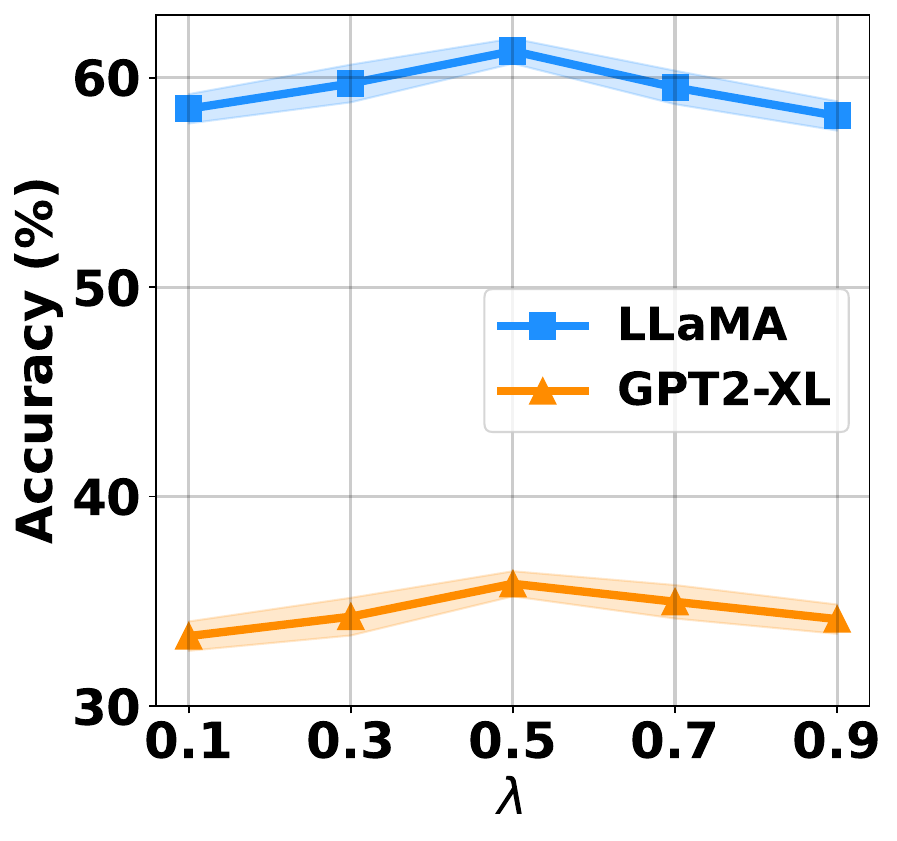}
\end{minipage}
\end{tabular}
\caption{The impact of the hyper-parameter $\lambda$ on EB (left) and GSM8K (right).}
\label{fig:hyper_lambda}
\end{figure}

We vary one hyper-parameter with others fixed. 
From Figure~\ref{fig:hyper_lambda}, as $\lambda$ increases,
the performance first increases and then drops, and it can achieve 
the best result when $\lambda=0.5$.

\section{Related Work}

\noindent\textbf{Multi-step Reasoning.} Reasoning has long been a key focus in natural language processing research. Initially, most studies concentrated on basic tasks like single-sentence language inference~\cite{zamansky2006natural,maccartney-manning-2009-extended, angeli-etal-2016-combining, hu-etal-2020-monalog, chen-etal-2021-neurallog} and commonsense inference~\cite{rajani-etal-2019-explain,latcinnik2020explaining,shwartz-etal-2020-unsupervised} in a single step. Lately, there has been a surge in research interest regarding multi-step reasoning and mathematical problem solving. The search space for a correct reasoning path is extremely vast and complex. Previous approaches~\cite{bostrom2022natural,creswell2022selection,ExplicitPlanning2023} predominantly emphasized a bottom-up reasoning approach, with a strong focus on system design. In contrast, our approach to constructing the cognitive tree adopts a top-down methodology, reducing the burden during the era of generative models. This top-down approach offers greater flexibility for application across models of varying scales.

\noindent\textbf{LLM as Evaluation.} The utilization of Language Models (LLMs) to evaluate the validity of their own predictions is gaining significance as a procedure in problem-solving. The introduction of the \textit{self-reflection} mechanism by~\citealp{Reflexion2023,Self-Refine2023,REFINER2023} involves LMs providing feedback to their generated candidates.
Tree of Thought~\cite{tot} and our approach share a common utilization of a tree-based structure for problem-solving. However, ToT primarily concentrates on tree construction and limited self-validation of intermediate results. According to the dual process theory, which suggests that validation requires deeper levels of thinking, our approach incorporates a contrastive learning method to enhance the model's ability to distinguish accurate results and facilitate comprehensive global validation of the generated outcomes.

\section{Conclusion}
In this paper, we proposed a new framework named CogTree to address complex logical reasoning and mathematical problems. The process of reasoning involves constructing a Cognitive Tree, where the nodes represent the decomposition of complex problems into sub-problems, and the edges represent judgments regarding the correctness of the decomposition. Based on the implementation of our approach over GPT2 and LLaMA, we have achieved results comparable to GPT-3.5 on EB and GSM8K datasets, indicating the effectiveness of our framework.

%\newpage

\section*{Limitations}
Due to the limitation of computational resources, we did not test our method on larger scale models. As the model size increases, using our approach may lead to further improvement in the accuracy of answers to these questions. Another direction worth exploring is diversifying the validation methods for the Reflective System, such as using multi-verification to compare the generated results and select the optimal answer.

\section*{Acknowledgements}
This work was supported in part by National Natural Science Foundation of China under Grant (No. 62072182, No. 92270119), the Fundamental Research Funds for the Central Universities, and by Alibaba Group through Alibaba Research Intern Program.

\appendix

\section{Appendix}
\label{sec:appendix}
Below we present some successful and failed cases of CogTree for analysis.

\begin{figure*}
\centering
\includegraphics[width=\textwidth]{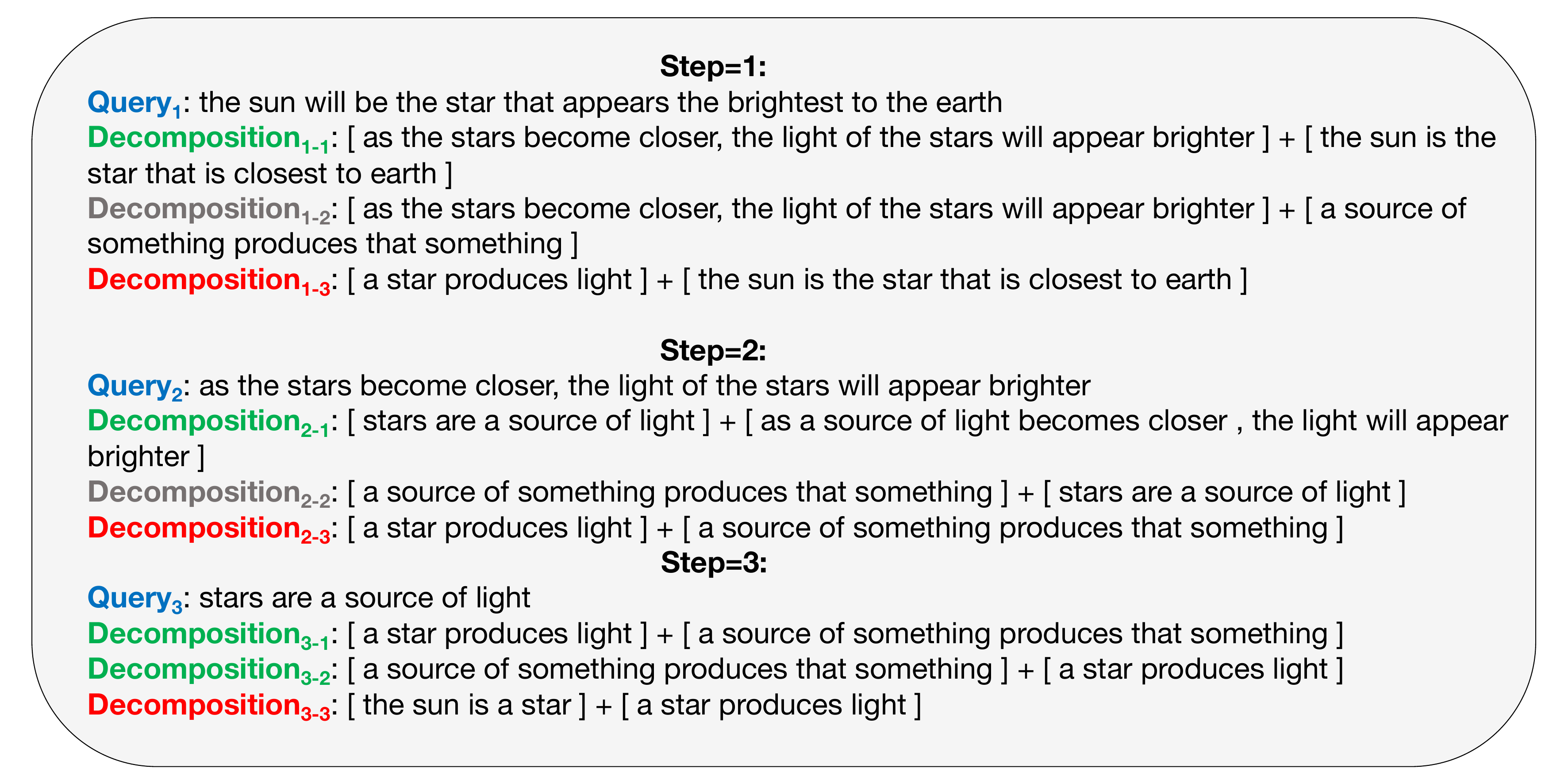}
\caption{A successful case in EB of 3 total steps. Green represents answers judged as ``sure'' by the Reflective System. Gray represents answers judged as ``likely'' by the Reflective System, and red represents answers judged as ``impossible'' by the Reflective System.}
\label{sucEB}
\end{figure*}

\begin{figure*}
\centering
\includegraphics[width=\textwidth]{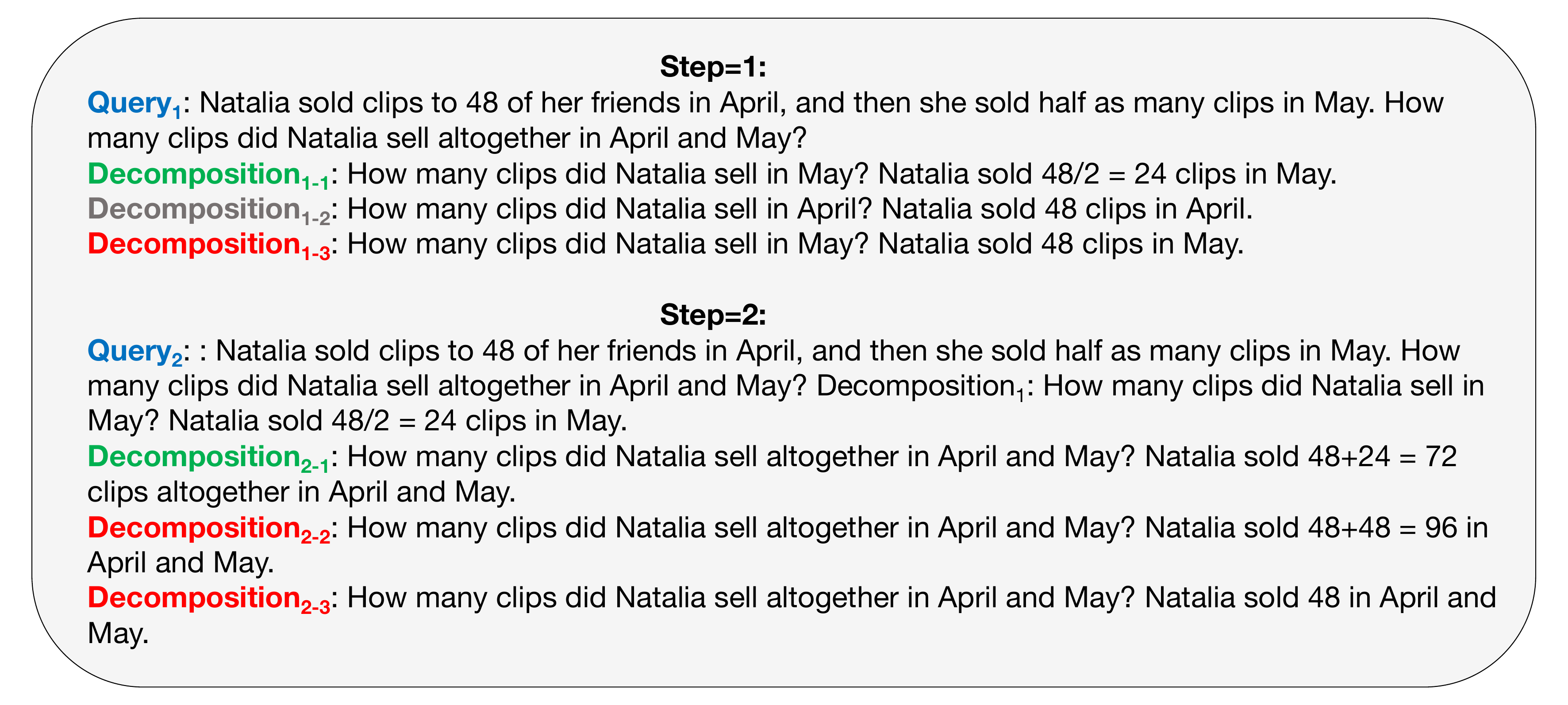}
\caption{A successful case in GSM8K of 2 total steps. Green represents answers judged as ``sure'' by the Reflective System. Gray represents answers judged as ``likely'' by the Reflective System, and red represents answers judged as ``impossible'' by the Reflective System.}
\label{sucGSM8K}
\end{figure*}

\begin{figure*}
\centering
\includegraphics[width=\textwidth]{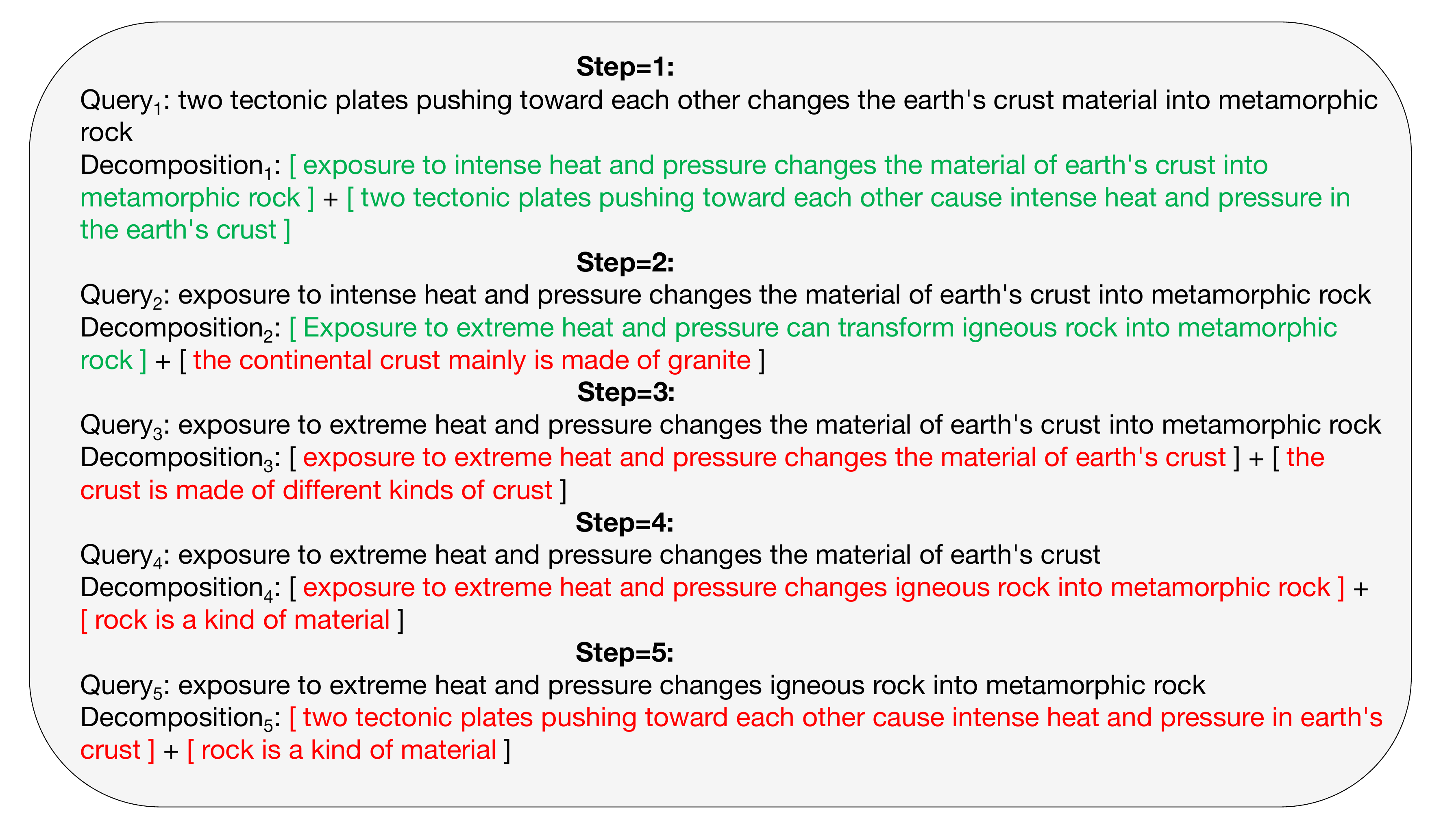}
\caption{A failed case in EB of 11 total steps. Our system stops generating after Step 5.}
\label{errorEB}
\end{figure*}

\begin{figure*}
\centering
\includegraphics[width=\textwidth]{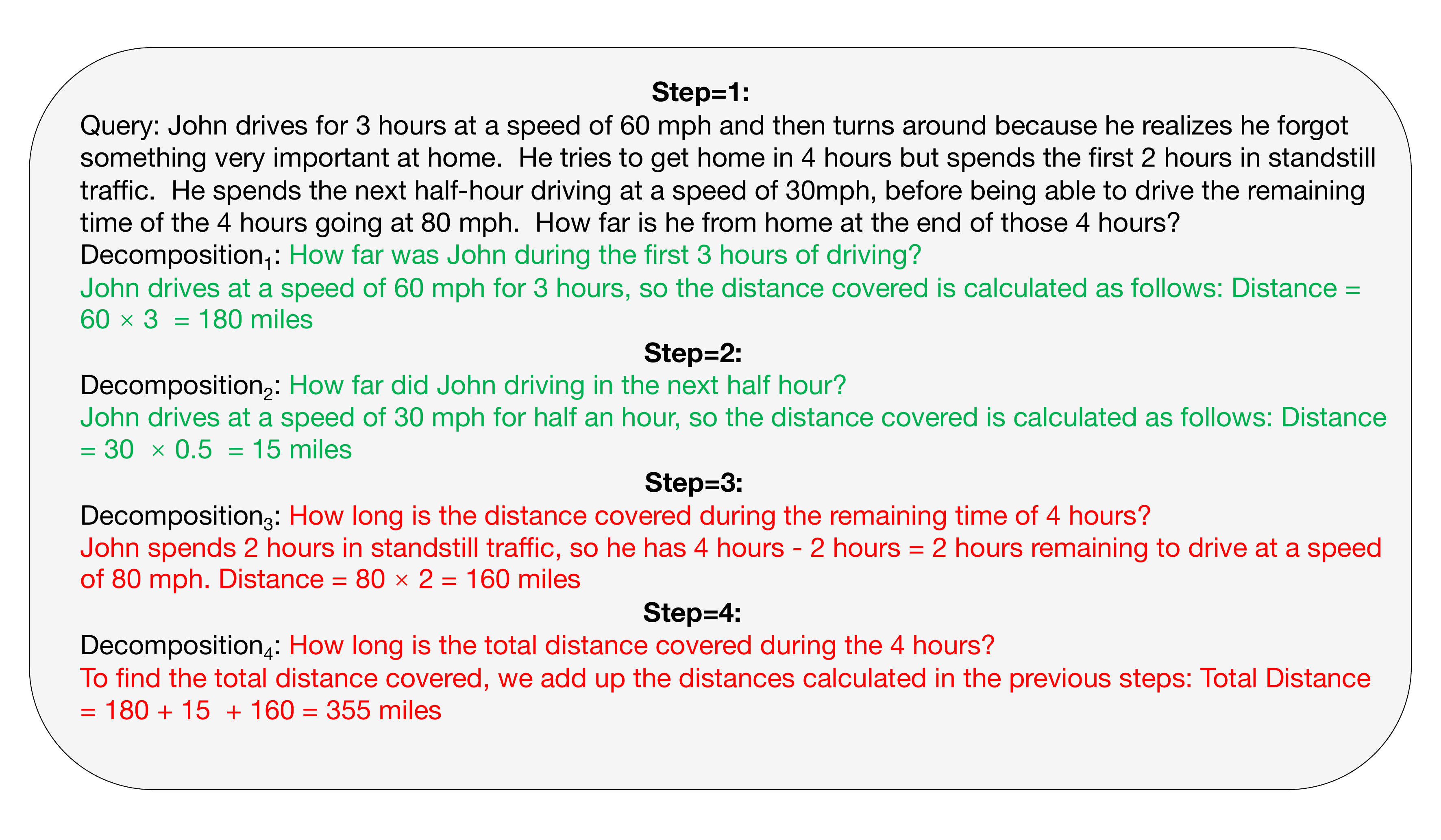}
\caption{A failed case in GSM8K of 6 total steps. Our system stops generating after Step 4.}
\label{errorGSM8K}
\end{figure*}

\end{document}